# Deterministic POMDPs Revisited

**Blai Bonet**
Departamento de Computación
Universidad Simón Bolívar
Caracas, Venezuela
bonet@ldc.usb.ve

## Abstract

We study a subclass of POMDPs, called Deterministic POMDPs, that is characterized by deterministic actions and observations. These models do not provide the same generality of POMDPs yet they capture a number of interesting and challenging problems, and permit more efficient algorithms. Indeed, some of the recent work in planning is built around such assumptions mainly by the quest of amenable models more expressive than the classical deterministic models. We provide results about the fundamental properties of Deterministic POMDPs, their relation with AND/OR search problems and algorithms, and their computational complexity.

## 1 Introduction

The simplest model for sequential decision making is the deterministic model with known initial and goal states. Solutions are sequences of actions that map the initial state into a goal state that can be computed with standard search algorithms. This model has been studied thoroughly in AI with important contributions such as A*, IDA*, and others [30, 34].

The deterministic model has strong limitations on the type of problems that can be represented: it is not possible to model situations where actions have non-deterministic outcomes or where states are not fully observable. In such cases, one must resort to more expressive formalisms such as Markov Decision Processes (MDPs) and Partially Observable MDPs (POMDPs). The generality of these models comes with a cost since the computation of solutions increase in complexity, specially for POMDPs, and thus one gains in generality but loses in the ability to solve problems. POMDPs, for example, are widely used as they offer one of the most general frameworks for sequential decision making [19], yet the known algorithms scale very poorly.

However, we have seen that an important collection of problems that involve uncertainty and partial information have a common characteristic: they have actions with deterministic outcomes and the observations generated at each decision stage also behave deterministically. Indeed, these models have been used in recent proposals for planning with incomplete information [15, 16, 27], appear in works of more general scope [15, 20] and about causation [31], and are used for learning partially-observable action models [1].

These models were briefly considered in Littman's thesis [21] under the name of Deterministic POMDPs (DET-POMDPs) for which some important theoretical results were obtained. Among others, he showed that a DET-POMDP can be mapped into an MDP with an exponential number of states and thus solved with standard MDP algorithms, and that optimal non-stationary policies of polynomial horizon can be computed in non-deterministic polynomial time. Unfortunately, DET-POMDPs briefly appeared as a curiosity of theoretical interest and then quickly fade out from consideration, to the point that, up to our knowledge, there are no publications on this subject neither from Littman or others.

Given the role of DET-POMDPs in recent investigations, motivated mainly by the quest of amenable models for decision making with uncertainty and partial information, we believe that DET-POMDPs should be further studied. In this paper, we carry out a systematic exploration of DET-POMDPs mainly from the complexity perspective yet we outline novel algorithms for them. We present three variants of the model: the fully observable, the unobservable and the general case, and two metrics of performance: worst- and expected-cost. As it will be shown, DET-POMDPs offer a tradeoff between the classical deterministic model and the general POMDP model. Furthermore, their characteristics permits the use of standard and novel AND/OR algo-rithms which are simpler and more efficient that the standard algorithms for POMDPs [32, 36, 37], or the transformation proposed by Littman.

The paper is organized as follows. First, we give examples of challenging problems that help us to establish the relevance of DET-POMDPs. We then present the definition and variants of the model in Sect. 3, the relation with AND/OR graphs and algorithms in Sect. 4, complexity analyses in Sect. 5, and finish with a brief discussion in Sect. 6.

## 2 Examples

Numerous DET-POMDPs problem had been used to evaluate and develop different algorithms for planning with uncertainty and partial information. For space reasons, we only provide few examples and brief descriptions for some of them.

**Mastermind.** There is a secret word of length $m$ over an alphabet of $n$ symbols. The goal is to discover the word by making guesses about it. Upon each guess, the number of exact matches and near matches is returned. The goal is to obtain a strategy to identify the secret.

**Navigation in Partially-Known Terrains.** There is a robot in a $n \times n$ grid that must navigate from an initial position to a goal position. Each cell of the grid is either traversable or untraversable. The robot has perfect sensing within its cell but the traversability of a subset of cells is unknown. The task is to obtain a strategy for guiding the robot to its destination [20].

**Diagnosis.** There are $n$ binary tests for finding out the state of a system among $m$ possible states. An instance consists of a $m \times n$ binary matrix $T$ such that $T_{ij} = 1$ iff test $j$ is positive when the state is $i$. The goal is to get a strategy for identifying the state [29].

**Coins.** There are $n$ coins from which one is a counterfeit of different weight, and there is a 2-pan balance scale. A strategy that spots the counterfeit and finds out whether is heavier or lighter is needed [30].

**Domains from IPC.** The problems in the 2006 and 2008 International Planning Competition for the track on conformant planning consisted of domains covering topics such as Blocksworld, circuit synthesis, universal traversal sequences, sorting networks, communication protocols and others [11], all of which are instances of DET-POMDPs.

## 3 Model and Variants

Formally, a DET-POMDP model is a tuple made of:

– a finite state space $S = \{1, \ldots, n\}$,

– finite sets of applicable actions $A_i \subseteq A$ for $i \in S$,

– a finite set of observations $O$,

– an initial subset of states $b_0 \subseteq S$, or alternatively an initial distribution of states $b_0 \in \Delta S$,

– a subset $T \subseteq S$ of goal (target) states,

– a *deterministic* transition function $f(i, a) \in S$, for $i \in S, a \in A_i$, that specifies the result of applying action $a$ on state $i$,

– a *deterministic* observation function $o(i, a) \in O$, for $i \in S, a \in A$, that specifies the observation received when *entering* state $i$ *after* the application of action $a$, and

– positive costs $c(i, a)$, for $i \in S, a \in A_i$ that tells the immediate cost of applying $a$ on $i$.

For simplicity, we assume that goal states are absorbing. That is, once a goal state is entered, the system remains there and incurs in no costs, hence $A_t = A$, $f(t, a) = t$ and $c(t, a) = 0$ for $t \in T$.

The two options for $b_0$ depend on whether the interest is in minimizing the worst-case total accumulated cost, or the expected total accumulated cost (see below). In any case, $b_0$ is called the initial *belief state*[1] and describes the different possibilities for the initial state which is not known a priori. Its importance is crucial since, under the assumption of deterministic transitions, if the initial state were known, then all future states would also be known, and the model reduces to the well-known deterministic model in AI [34]. Hence, the only source of uncertainty in DET-POMDP models comes from the initial state which further induces an uncertainty on the observations generated at each decision stage. Nonetheless, the model remains challenging, with respect to expressivity and computation, as exemplified in the previous section.

Although the state of the system may not be fully observable, it is possible (and indeed useful) to consider preconditions for actions. The role of preconditions is to permit the knowledge engineer to design economical representations by leaving out from the specification unimportant details or undesirable situations. For example, if one does not want to model the effects of plugging a 120V artifact into a 240V outlet, then a simple precondition can be used to avoid such situations. Preconditions in the DET-POMDP model are expressed through the sets of applicable actions $A_i$. As it is standard in planning, situations in which there are no actions available are called dead ends.

---

[1]The term 'belief state' refers to a subset of states (or a probability distribution on states) that is deemed possible by the agent at a given point in time.

### 3.1 Optimality Criteria

Our goal is to compute strategies that permit the agent to act *optimally*. We consider two optimality criteria, and three variants of the model. For optimality, we consider the **minmax** that minimizes the worst-case cost of a policy, and the **minexp** that minimizes the expected cost of a policy. The variants of the model depend on the observation model:

- unobservable models in which $O$ is a singleton and thus observations return no information about the state. This class is a subclass of the so-called conformant problems in planning [13].

- fully-observable models in which $o(i, a) = i$. This class corresponds to fully observable MDPs except that the initial state is unknown, but after the application of the first action the state is revealed.

- models with no assumptions on the observations. This case corresponds to general DET-POMDPs and include the previous cases.

Two optimization criteria and three variants combine into six models: unobservable models with the minmax and minexp criteria, fully-observable models with the minmax and minexp criteria, and general models with minmax and minexp criteria. Among these, the unobservable and the general models are the interesting ones; fully-observable models are trivial and pathological and thus will not be considered again.

For partially-observable problems, the most general form of a policy is a function that maps belief states into actions. Beliefs are subsets of states for minmax models and probability distributions over states for minexp models. However, an optimal policy does not need to specify an action for all possible beliefs: it just need to consider the beliefs that could appear during the execution of the policy.

### 3.2 Belief Dynamics and Closed Policies

In order to execute an action, the agent must be certain about its applicability no matter what is the current state of the environment. Otherwise, the model might end up in an inconsistent configuration with respect to the specification. Hence, the set $A_b$ of applicable actions at $b$ is $A_b \doteq \cap \{A_i : i \in \sup(b)\}$ where $\sup(b)$, called the *support* of $b$, is the collection of states that are deemed possible at $b$, i.e. $\sup(b) \doteq b$ for subsets of states and $\sup(b) \doteq \{i : b(i) > 0\}$ for distributions on states. Once an applicable action is executed, each state in $\sup(b)$ changes into a new state making up a new belief $b_a \doteq \{f(i, a) : i \in \sup(b)\}$ for subsets, and $b_a(j) \doteq \sum_{f(i,a)=j} b(i)$ for distributions.

On the real environment though, the current state transforms into a new state and an observation is generated, which is used to *filter out* states inconsistent with it; i.e. the filtered belief is $b_a^o \doteq \{i \in b_a : o(i, a) = o\}$ for subsets, and, for distributions, is

$$b_a^o(i) \doteq \begin{cases} 0 & \text{if } b_a(i) = 0 \text{ or } o(i, a) \neq o, \\ b_a(i)/b_a(o) & \text{otherwise,} \end{cases}$$

where $b_a(o)$ is a normalization constant.

From the point of view of the agent, which does not know the current state of the environment, the set of observations that may appear after the application of $a$ at $b$ is $o(b, a) \doteq \{o(i, a) : i \in \sup(b_a)\}$. Under the minexp criterion, each observation has probability $b_a(o)$ of being received once action $a$ is applied at $b$.

Littman proposes an interesting representation for the reachable beliefs in the probabilistic case [21], in which a reachable belief $b$ is represented by a table $t_b : S \to S \cup \{\bot\}$ as follows. For $b_0$, $t_{b_0}(i)$ is $i$ or $\bot$ whether $b_0(i) > 0$ or not, and for $b_a^o$ where $b$ is a reachable belief, $a \in A_b$ and $o \in o(b, a)$, the table $t_{b_a^o}$ is given by $t_{b_a^o}(i) = \bot$ if $t_b(i) = \bot$, $b_a(i) = 0$ or $o(i, a) \neq o$, and $t_{b_a^o}(i) = f(i, a)$ otherwise. In words, the entry $t_b(i)$ tells what would be the current state had the initial state been equal to $i$. The relation between $b$ and $t_b$ is

$$b(i) \propto \sum_{j \in S} b_0(j) |\{j : t_b(j) = i\}|.$$

We are interested in optimal policies that are *closed with respect to the initial belief* $b_0$. Namely, that the policy is defined over all the beliefs that may appear during the execution of the policy. We say that a policy $\pi$ is closed in a subset $B$ of beliefs if $\pi$ is defined on each $b \in B$, and $b_{\pi(b)}^o \in B$ for each $b \in B$ and $o \in o(b, \pi(b))$, and that $\pi$ is closed with respect to $b_0$, or just closed, if $b_0 \in \text{Dom}(\pi)$ and $\pi$ is closed in $\text{Dom}(\pi)$. We will only consider closed policies. The set of reachable beliefs from $b_0$ using (closed) policy $\pi$ is the minimum subset $R \doteq R(b_0, \pi) \subseteq \text{Dom}(\pi)$ such that $b_0 \in R$ and $\pi|_R$ is closed with respect to $b_0$.[2] A policy $\pi$ is *minimal* if $\pi = \pi|_{R(b_0, \pi)}$.

### 3.3 Graphs, Costs and Optimal Policies

Let us reason about the trajectories generated by a policy $\pi$. At the beginning, the agent applies the action $a_0 = \pi(b_0)$. Then, the agent incurs in certain immediate cost, which depends on the current state and $a_0$, and receives an observation $o_0$, which also depends on the current state and action, that is used by the agent to update its initial belief into $b_1 = (b_0)_{a_0}^{o_0}$. At the second decision stage, the agent applies the action $a_1 = \pi(b_1)$, incurs in a cost and receives a new observations $o_1$ that is used to update the belief into $b_2 = (b_1)_{a_1}^{o_1}$, and so on. This process continues until the agent is certain that a goal state has been reached, i.e. until the current belief $b$ is a target belief characterized by $\sup(b) \subseteq T$. It is important to remark that during the execution of a policy the agent cannot predict with certainty neither the costs nor the observations since s/he is not certain about the current state of the environment.

---

[2] $f|_A$ means the restriction of $f$ to $\text{Dom}(f) \cap A$.

All possibles trajectories of $\pi$ seeded at $b_0$ form a labeled directed (multi-)graph $G_\pi = (V, E, \ell)$ where $V = R(b_0, \pi)$ and there is an edge $(b, b')$ labeled with $o$ iff $b$ is non-target and $b' = b_{\pi(b)}^o$ for some $o \in o(b, \pi(b))$. In probabilistic models, edges $(b, b')$ labeled with $o$ have attached probabilities equal to $b_{\pi(b)}(o)$.

Deterministic actions imply $|\sup(b_a)| \leq |\sup(b)|$ with strict inequality only if $f(i,a) = f(j,a)$ for some $i \neq j$ in $\sup(b)$. Therefore, if $a$ is applicable in $b$ and $\{b_1, \ldots, b_m\}$ is the collection of possible beliefs after observations, then the collection $\{\sup(b_i)\}_{i=1}^m$ is a partition of $\sup(b_a)$ and thus $\sum_{o \in o(b,a)} |\sup(b_a^o)| \leq |\sup(b)|$ for each $a \in A_b$. Furthermore, if $b <_\pi b'$ denotes the existence of a path $b \leadsto b'$ in $G_\pi$ and $b <_\pi^d b'$ the existence of a *deterministic* path, i.e. one in which each node has outdegree of one, then

**Theorem 1.** *Let $G_\pi$ be the graph for $\pi$. Then, (a) if $b <_\pi b'$ then $|sup(b)| \geq |sup(b')|$, (b) if $b <_\pi b'$ and $|sup(b)| = |sup(b')|$ then $b <_\pi^d b'$, and (c) for minexp, if $b <_\pi b'$ and $sup(b) = sup(b')$ then $b(i) = b'(\sigma_i)$ where $\sigma$ is a permutation of $sup(b)$.*

An immediate consequence is that for minexp models the set of reachable beliefs is finite *independently of the initial belief*. This fact was also observed by Littman who gave the upper bound $(1+|S|)^{|S|}$ on the maximum number of reachable beliefs as this is the maximum number of tables. This is a marked difference with respect to standard POMDPs in which the number of reachable beliefs is typically infinite as well as the size of the policy graphs.

**Theorem 2.** *The set $R(b_0, A)$ of reachable beliefs is finite, the graph $G_\pi$ is finite for each policy $\pi$, and the number of minimal policies is also finite.*

To define the cost of a policy, we consider the cases of whether $G_\pi$ is a DAG or not. In the latter case, there is a cycle in $G_\pi$ along a deterministic path and thus, for both optimization criteria, $\pi$ incurs in infinite cost once it gets trapped into the cycle.[3] Therefore, if $G_\pi$ has a cycle, its cost at $b_0$ is set to infinity; i.e. $V_\pi^{\max}(b_0) = V_\pi^{\exp}(b_0) \doteq \infty$.

---
[3]In MDPs, it is customary to use a factor to discount future costs at a geometric rate so that every policy has finite cost. We do not consider such factors as often they are difficult to interpret and justify.

If $G_\pi$ is a DAG, the cost assigned by $\pi$ at each belief is defined inductively bottom-up from the sink nodes up to the source $b_0$. Indeed, being $\pi$ closed, the sinks are target beliefs which has zero cost under both criteria, i.e. $V_\pi^{\max}(b) = V_\pi^{\exp}(b) \doteq 0$ if $\sup(b) \subseteq T$. Once all successors of $b$ get values, the value at $b$ is defined as

$$V_\pi^{\max}(b) \doteq c^{\max}(b, \pi(b)) + \max_{(b,b') \in G_\pi} V_\pi^{\max}(b'),$$

$$V_\pi^{\exp}(b) \doteq c^{\exp}(b, \pi(b)) + \sum_{o \in o(b,\pi(b))} b_a(o) \cdot V_\pi^{\exp}(b_a^o),$$

where the costs over beliefs are defined as $c^{\max}(b,a) \doteq \max_{i \in b} c(i,a)$ and $c^{\exp}(b,a) = \sum_{i \in S} b(i) \cdot c(i,a)$. The cost assigned by policy $\pi$ to belief $b_0$, either finite or infinite, is called the cost of $\pi$. Clearly, if $G_\pi$ is a DAG, then $V_\pi^{\exp}(b_0) \leq V_\pi^{\max}(b_0) < \infty$.

A policy $\pi$ is preferred to policy $\pi'$ under optimization criterion $\Delta$ if $V_\pi^\Delta(b_0) < V_{\pi'}^\Delta(b_0)$. A policy $\pi$ is optimal under $\Delta$ if no other policy is preferred to it. Hence, by Theorem 2, if there is a policy of finite cost, then there is an optimal policy $\pi^*$ whose graph is a DAG. From now on, when we say an optimal policy, we mean an optimal policy of finite cost. If there is no policy of finite cost, we say there is no optimal policy.

### 3.4 Policy Forms and Sizes

**Unobservable Models.** As there is no information on which to branch, plans for unobservable models are linear sequences of actions that take the initial belief into a goal belief. Hence, $G_\pi$ is a single path of form

$$b_0 \rightarrow b_1 \rightarrow b_2 \rightarrow \cdots \rightarrow b_{n-1} \rightarrow b_n.$$

By Theorem 1, $|\sup(b_0)| \geq \cdots \geq 1$. Let us say that a "jump" occurs at $b_i$ if $|\sup(b_i)| > |\sup(b_{i+1})|$, and that $\langle b_i, \ldots, b_{i+m} \rangle$ is a "chunk" if it is a maximal subsequence with $|\sup(b_i)| = \cdots = |\sup(b_{i+m})|$. There are at most $|S|$ jumps in the sequence so we need to bound the size of a largest chunk. Theorem 1 implies that the actions in the chunk map states in a 1-1 way, i.e. that actions behave like permutations over states. As it will be shown later, the size of a chunk can be exponential in some problems.

Let us introduce the notion of *diameter of a model* that allow us to bound the length of a largest chunk. For a set of actions $A$ and belief $b$, denote with $R(b, A)$ the set of beliefs reachable from $b$ using only actions in $A$, and with $R^*(b, A)$ the set of beliefs with supports of size $|\sup(b)|$ that are reachable from $b$ using only actions in $A$. A belief in $R^*(b, A)$ is said to be *k-reachable* if it can be reached from $b$ through the application of at most $k$ actions from $A$. The *diameter* of $R^*(b, A)$ is the least integer $k$ such that every $b' \in R^*(b, A)$ is $k$-reachable, and the diameter of a model $M$ is the maximum over the diameters of $R^*(b, A)$ for all $b \in R(b_0, A)$. The model has polynomial diameter if its diameter is $O(poly(|S|, |A|))$.

If $R(b_0, A)$ is of polynomial size, the model is of polynomial diameter but the converse is not necessarily true. If $M$ is of polynomial diameter, the lengths of the chunks are of polynomial length. Thus, unobservable problems of polynomial diameter have optimal policies of polynomial size.

**General Models.** Optimal policies are DAGs in which paths correspond to sequences of beliefs with non-increasing supports. As before,

**Theorem 3.** *Models of polynomial diameter have optimal policies of polynomial size.*

*Proof.* It remains to show the claim for general models. Let $\pi$ be an optimal policy. A subset $B$ of beliefs is called an *anti-chain* iff there are no two different beliefs $b, b' \in B$ with $b <_\pi b'$. An induction shows that if $B$ is an anti-chain, then $|\sup(b_0)| \geq \sum_{b \in B} |\sup(b)|$. Thus, since $|\sup(b_0)| \leq |S|$, all anti-chains have size at most $|S|$. On the other hand, the assumption implies that all paths in $G_\pi$ have polynomial length. Therefore, $G_\pi$ is of polynomial size since the beliefs at depth $k$ form an anti-chain and thus their number is at most $|S|$. □

## 4 AND/OR Graphs

We establish a relation between policies and AND/OR graphs that can be exploited by algorithms. An AND/OR graph is a tuple $G = (V_1 \cup V_2, E, T, n_0, c)$ where $V_1$ and $V_2$ are finite sets of AND and OR nodes, $E \subseteq (V_1 \cup V2) \setminus T \times (V_1 \cup V_2)$ is a subset of directed edges between nodes, $T \subseteq V_2$ is subset of terminal nodes, $n_0 \in V_1 \cup V_2$ is the root node, and $c : E \cup T \to \mathbb{R}^*$ is a cost function over edges and terminal nodes [26]. A solution is a subgraph $H$ spanned by a subset of edges $H(E)$ such that (1) $n_0 \in H$, (2) if $n \in H \setminus T$ is AND node then all its outbound edges are in $H(E)$, and (3) if $n \in H \setminus T$ is OR node then one of its outbound edges is in $H(E)$. A solution is valid iff it is acyclic. Costs $c(H)$ can be associated to valid solutions $H$ inductively from the sinks up to the root by

$$V_H(n) \doteq \begin{cases} c(n) & n \in T \\ \max_{(n,n') \in H(E)} c(n, n') + V_H(n') & n \in V_1 \\ c(n, n') + V_H(n') & n \in V_2 \end{cases}$$

A stochastic graph associates probabilities $p(n, n')$ to the edges $(n, n')$ incident at AND nodes $n$ so that $\sum_{(n,n') \in E} p(n, n') = 1$. The cost $V_H$ is similarly defined except that $V_H(n) \doteq \sum_{(n,n') \in H(E)} (c(n, n') + V_H(n'))p(n, n')$ for AND nodes. In any case, the cost $c(H)$ is defined as $V_H(n_0)$, and a solution is optimal if its cost is no larger than the cost of any other solution.

### 4.1 From Models to Graphs and Algorithms

The relation between DET-POMDPs and AND/OR graphs is direct. For a model $M$, we construct an AND/OR graph $G_M$ such that the solutions of $G_M$ coincide with the solutions of $M$. Indeed, given a model $M$ with minmax criterion, the graph $G_M$ is given by

- $V_1 \doteq \{b_a : b \in R(b_0, A), a \in A_b\}$,
- $V_2 \doteq R(b_0, A)$,
- $T \doteq \{b \in V_2 : \sup(b) \subseteq T\}$,
- if $b \in V_2 \setminus T$ then $(b, b_a) \in E$ for $a \in A_b$,
- if $b_a \in V_1$ then $(b_a, b_a^o) \in E$ for $o \in o(b, a)$,
- $n_0 \doteq b_0$,
- $c(b) \doteq 0$ for $b \in T$,
- $c(b, b_a) \doteq c^{\max}(b, a)$ for $b \in V_2$ and $a \in A_b$,
- $c(b_a, b_a^o) \doteq 0$ for $b_a \in V_1$ and $o \in o(b, a)$.

For minexp, the graph must be extended into a stochastic graph with labels $p(b_a, b_a^o) = b_a(o)$, and $c(b, b_a)$ must be replaced by $c^{\exp}(b, a)$. Also observe that the beliefs in the graph can be represented with probability distributions or with Littman's tables.

This relation permits the use of diverse algorithms. Firstly, since the number of reachable beliefs is finite, then all reachable beliefs can be enumerated and Value or Policy Iteration applied on the resulting MDP over belief space (this is essentially Littman's proposal); both algorithms are guaranteed to converge in a finite number of steps. However, we can do better since optimal policies are acyclic and thus *consistently improving policies* exist. Therefore, more efficient algorithms can be used such as label setting methods and Dial's algorithm [3, 7].

Explicit algorithms as the above require enough memory to compute the set of reachable beliefs. If no such memory is available, search-based algorithms that generate beliefs incrementally and that use admissible heuristics to focus the search must be used. Relevant algorithms in this class are the classical AO* [6, 23, 26], LAO* [14], RTDP-like algorithms [2, 4], and LDFS [5]. AO* can only be used on acyclic graphs; LAO*, RTDP and LDFS do not have this restriction.

## 5 Complexity

The complexity of POMDPs had been thoroughly studied. Results for different optimization criteria, probabilistic and non-deterministic variants, and so on are known [17, 22, 25, 28, 33]. Littman [21] obtained important complexity results about DET-POMDPs for problems with non-negative costs (i.e. with non-positive rewards):

- deciding the existence of a policy that incurs in zero cost for an infinite-horizon DET-POMDP is PSPACE-complete, and
- deciding the existence of a policy of polynomial length that incurs in zero cost for a DET-POMDP is NP-complete

In this section, we extend this results as follows:

- deciding the existence of a policy of finite cost for DET-POMDPs of polynomial diameter is NP-complete,
- there are DET-POMDPs on which all policies of finite cost are of super-polynomial size,
- give new class of problems on which the existence of policies can be checked in non-deterministic polynomial time, and
- give some sufficient conditions for testing whether a DET-POMDP is of polynomial diameter.

All these results, as well as Littman's results, are based in the assumption that the models are encoded in a flat language; that is, that the models are encoded with $O(poly(|S|, |A|, h))$ bits where $h$ is the maximum number of bits needed to specify a cost or probability. The general question that we address here is whether there is a policy that reaches the goal with certainty. Namely,

POLICY EXISTENCE FOR FLAT MODELS (PEF): Is there a policy of finite cost for flat model $M$?

The fact that PEF only considers flat models is important because with compact representations, an exponential number of states can be described and thus a double-exponential number of beliefs could be needed.

**Theorem 4.** *The PEF problem for unobservable models of polynomial diameter is* NP-complete.

*Proof.* Inclusion is direct by guessing a polynomially-sized policy and checking it. For hardness, we reduce SAT using the method in [25] almost verbatim. Let $\phi$ be a 3-CNF formula with variables $x_1, \ldots, x_n$ and clauses $C_1, \ldots, C_m$. A variable $x$ 1-appears (0-appears) in clause $C$ if $x \in C$ ($\overline{x} \in C$). We construct an unobservable model $M(\phi)$ as follows. The set of states is $S = \{[x_i, C_j] : 1 \leq i \leq n, 1 \leq j \leq m\} \cup \{\mathsf{t}, \mathsf{f}\}$. States $\mathsf{t}$ and $\mathsf{f}$ mean satisfiable and unsatisfiable respectively. The actions $set(i, v)$, for $1 \leq i \leq n$ and $v \in \{0, 1\}$, are used to set the value of variables: $set(i, v)$ maps state $[x_i, C_j]$ ($[x_n, C_j]$) to either $\mathsf{t}$ or $[x_{i+1}, C_j]$ (to either $\mathsf{t}$ or $\mathsf{f}$) whether $x_i$ $v$-appears in $C_j$ or not. The initial belief is $b_0 = \{[x_1, C_j] : 1 \leq j \leq m\}$ and $\mathsf{t}$ is the unique target state. $M(\phi)$ is of polynomial size and diameter, and has a policy of finite cost iff $\phi$ is satisfiable. □

**Corollary 5.** *The PEF problem for general models of polynomial diameter is* NP-complete.

*Proof.* Hardness is direct since unobservable models are a special case. Inclusion follows by guessing a policy of polynomial size (Theorem 3) and checking it. □

Sometimes, it is easy to verify that a model is of polynomial diameter. Indeed, if $|sup(b_0)| = k$ the number of reachable beliefs is $O(n^k)$. If $k$ is a constant then the number of reachable beliefs is polynomial as well as the diameter. For another case, consider the (global) transition graph $T_M$ that is a directed graph whose nodes are the states and there is an edge $(i, j)$ iff there is action $a \in A_i$ such that $j = f(i, a)$. If $T_M$ is acyclic, except for self-loops, the model has linear diameter. Examples are knowledge-gathering problems such as Mastermind, 12-coins and Diagnosis. The IPC problem for sorting networks is also of this type.

**Theorem 6.** *If $|sup(b_0)|$ is constant, then the diameter is polynomial. If $T_M$ is acyclic, except perhaps for self-loops, then the diameter is linear.*

As noted in Theorem 1, there is a close connection between the diameter of a model and permutations over states. Thus, the study of permutations provides important insight for estimating the diameter of models.

### 5.1 Permutation Groups and Their Diameter

The set of permutations with composition forms a multiplicative group. The order of a permutation $\sigma$ is the least $n$ such that $\sigma^n$ is the identity permutation. It is well known that a permutation can be written in cycle notation as a product of disjoint cycles, and that the order of $\sigma = C_1 C_2 \ldots C_m$ (in cycle notation) is the least common multiple for the lengths of the cycles. For example, $\sigma = \begin{pmatrix} 1 & 2 & 3 & 4 & 5 & 6 & 7 & 8 \\ 3 & 4 & 5 & 7 & 8 & 6 & 2 & 1 \end{pmatrix}$ can be written as $\sigma = (1, 3, 5, 8)(2, 4, 7)(6)$ and its order is $ord(\sigma) = \text{lcm}\{4, 3, 1\} = 12$.

Consider now an unobservable model with a single action $a$ with empty precondition that corresponds to a permutation $\sigma = C_1 C_2 \cdots C_m$ over states. Furthermore, suppose that $b_0$ has $m$ states one from each cycle $C_i$. Then, the repeated application of $a$ over $b_0$ generates a trajectory $\langle b_0, \ldots, b_k \rangle$ of *different* beliefs if $k < ord(\sigma)$. Since observations filter nothing, $|sup(b_0)| = \cdots = |sup(b_k)|$ and there is a chunk of length $k + 1$. If $k$ is large, the size of an optimal policy could be large as well. We use this idea to proof

**Theorem 7.** *There is an unobservable model for which all policies are of super-polynomial size.*

*Proof.* Let $S = \{1, \ldots, n\}$. First, we show that there is a permutation $\sigma$ over $S$ with super-polynomial order. If $\sigma = C_1 \ldots C_m$ in cycle notation, then $|C_1| + \cdots + |C_m| = n$ and $ord(\sigma) = \text{lcm}\{|C_1|, \ldots, |C_m|\}$, so we need to show that there are integers $\{d_1, \ldots, d_m\}$ whose sum is $n$ and lcm is super-polynomial in $n$.

The Prime Number Theorem says that the number of primes less than $N$ is asymptotically equal to $N/\log N$. Hence, the number of primes in $[n^{1/2}, n^{3/4}]$ is asymptotically equal to $4n^{3/4}/3\log n - 2n^{1/2}/\log n$ which dominates $n^{1/4}$. Therefore, for sufficiently large $n$, we can choose $\lfloor n^{1/4} \rfloor$ *different* primes in $[n^{1/2}, n^{3/4}]$. Their sum is bounded by $n^{3/4} \times n^{1/4} = n$ and thus these primes can be extended into a collection $\{d_1, \ldots, d_m\}$ of integers whose sum is equal to $n$. Since all primes are different and at least $n^{1/2}$, their lcm is at least $n^{\frac{1}{2}n^{1/4}}$ which dominates $n^k$ for any constant $k$. Estimations on the expected order of random permutations are also known [9, 12, 38].

For the model, let $a$ be an action that is like $\sigma$ and let $b_0$ be the subset containing the "first" state from each cycle $C_i$. Let $a'$ be another action that maps $s$ into a target if $s$ is the "last" state in a cycle, or into $s$ otherwise. The optimal plan is the sequence of $ord(\sigma) - 1$ repetitions of $a$ followed by $a'$. □

However, even in cases of non-polynomial diameter, we can test in non-deterministic polynomial-time the existence of policies for some models (proof below).

**Theorem 8.** *If each action is a permutation with empty precondition, then the PEF problem is in NP, even if the model is not of polynomial diameter.*

The group $S_n$ of all permutations over $n$ elements is called the symmetric group of degree $n$. Let $A \subseteq S_n$ be a subset. Then, $A$ *generates* the subgroup $\langle A \rangle$ of all permutations that can be formed by finite compositions of elements from $A$. A permutation $\sigma \in \langle A \rangle$ is $k$-expressible if it is the product of at most $k$ permutations from $A$. The diameter of $\langle A \rangle$ is the least integer $k$ such that every permutation in $\langle A \rangle$ is $k$-expressible. Problems related to generated groups have a tradition in computer science. For example, Furst, Hopcroft and Luks [10] gave a polynomial-time algorithm for deciding whether a permutation $\sigma$ is generated by the set $A$, and Jerrum [18] showed that computing the size of a smallest generator set is PSPACE-complete.

*Proof Sketch for Theorem 8.* We do it for unobservable models; the proof for general models is a bit more involved. Since actions are permutations, the size of beliefs do not change with actions. If there is a solution $b_0 \to \cdots \to b_n$ where $b_n$ is a target belief, then by Theorem 1, there is a permutation $\sigma$ such that $sup(b_0)$ is mapped 1-1 into $sup(b_n)$. Thus, it is enough to guess $\sigma$ and then test, in polynomial-time using the algorithm in [10], if $\sigma$ is generated by the actions. □

More important to us are results about diameters of groups. Driscoll and Furst [8] showed that if the generators are all cycles of bounded length, the diameter is $O(n^2)$, while McKenzie [24] showed that if all generators move at most $k$ elements, the diameter is $O(n^k)$.

**Theorem 9.** *If every action has empty precondition, and is a permutation that moves a constant number of states or all cycles are of bounded length, then the model is of polynomial diameter.*

## 6 Discussion

We have shown that the Deterministic POMDP model first studied by Littman is more relevant than previously thought for two reasons. First, a number of important and challenging problems in planning correspond to such models, and second since they provide a tradeoff between the restricted yet efficient classical deterministic model and the general but inefficient POMDP model. We have shown novel complexity results that show this tradeoff: while checking the existence of plans in classical deterministic (flat) problems is NL-complete (via $st$-REACHABILITY [35]), and checking the existence of plans for POMDPs is EXPTIME-complete [21], we have that checking the existence of plans for DET-POMDPs of polynomial diameter is NP-complete, and that for DET-POMDPs in which the actions are permutations with empty precondition is in NP. Furthermore, we give polynomial-time checkable conditions for polynomial diameter.

We also proposed a relation between DET-POMDPs and AND/OR graphs that permit the use of general AND/OR search algorithms. Although we did not perform an experimental evaluation, it is clear that such algorithms should outperform in practice the explicit mapping of a DET-POMDP into an MDP.

It is important to emphasize that DET-POMDPs are not "general" POMDPs, and hence one should be careful when evaluating general algorithms with them. Due to the reduced complexity of the model, DET-POMDPs must be tackled as a more specialized class which is likely to scale better.

In the future, we would like to further study the relation between DET-POMDPs and permutation groups and to implement algorithms based of AND/OR search.

**Acknowledgments.** Thanks to the reviewers for valuable comments and pointers to related work.